# Transfer Learning based Detection of Diabetic Retinopathy from Small Dataset

Misgina Tsighe Hagos
M.Tech Student
Department of Computer Science and Engineering
Sharda University
tsighemisgina@gmail.com

Shri Kant
Research and Technology Development Center
Department of Computer Science and Engineering
Sharda University
shri.kant@sharda.ac.in

**Abstract**

Annotated training data insufficiency remains to be one of the challenges of applying deep learning in medical data classification problems. Transfer learning from an already trained deep convolutional network can be used to reduce the cost of training from scratch and to train with small training data for deep learning. This raises the question of whether we can use transfer learning to overcome the training data insufficiency problem in deep learning based medical data classifications. Deep convolutional networks have been achieving high performance results on the ImageNet Large Scale Visual Recognition Competition (ILSVRC) image classification challenge. One example is the Inception-V3 model that was the first runner up on the ILSVRC 2015 challenge. Inception modules that help to extract different sized features of input images in one level of convolution are the unique features of the Inception-V3. In this work, we have used a pre-trained Inception-V3 model to take advantage of its Inception modules for Diabetic Retinopathy detection. In order to tackle the labelled data insufficiency problem, we subsampled a smaller version of the Kaggle Diabetic Retinopathy classification challenge dataset for model training, and tested the model's accuracy on a previously unseen data subset. Our technique could be used in other deep learning based medical image classification problems facing the challenge of labeled training data insufficiency.

## 1. Introduction
### 1.1 Diabetic Retinopathy

It was reported that 422 million people were living with diabetes mellitus in 2014 [1]. Over one third of the individuals that were living with diabetes in 2010 have shown signs of Diabetic Retinopathy (DR), and a third of those with signs of DR were affected by vision threatening DR. In 2050, it is expected that the number of Americans 40 years or older with DR will triple from 6.7 million in 2005 to 19.4 million; and among those 65 years of age or older Americans, it will increase from 3 million to 11.8 million [2]. All individuals with diabetes mellitus will eventually be at risk of developing DR [3]. DR is one of the leading causes of vision impairment and blindness.

DR is a complication of both type 1 and type 2 diabetes. It is caused when diabetes damages the blood vessels in the retina. DR is categorized into 3 categories: non-proliferative DR (NPDR), proliferative DR (PDR) and diabetic macular edema (DME). NPDR is characterized by visible features such as microaneurysms, hemorrhages and intraretinal microvascular abnormalities. PDR is the advanced stage of NPDR, and it is characterized by the production of new retinal blood vessels and retinal detachment may occur in this category. DME occurs when the blood-retina barrier (BRB) breaks. And it can be seen across all severity stages of NPDR and PDR. Break down of BRB leads to fluid leakage into the neural retina, which in turn causes abnormal retinal thickening [4]. PDR and DME are sight threatening categories of DR.

Wilkinson *et al.* [5] reached a consensus to classify DR into 5 stages according to disease severity as shown in Figure 1. The first stage is 'no apparent retinopathy' where there is no damage caused by diabetes mellitus to the retina. Few microaneurysms start to appear in the second stage which is labelled as 'mild DR'. 'Moderate DR' is the third stage, which is characterized by multiple microaneurysms, dot and blot hemorrhages and/or cotton wool spots. The fourth stage, called 'severe' DR is characterized by intraretinal microvascular abnormalities, venous beading and cotton wool spots. 'Proliferative DR' is the final stage of DR in which new retinal blood vessels start getting produced, blood leaks into the vitreous humor of the eye, and retinal detachment may start to appear. An ophthalmologist can diagnose DR by using different imaging techniques such as fundus photography, optical coherence tomography, and fluorescein angiography [6]. In this work, we have used fundus images that are collected by using color fundus photography.

It has been seen that longer diabetes duration, bad blood pressure and glycemic control increase the risk of DR [3]. In the Wisconsin Epidemiologic Study of DR (WESDR), the incidence of DR (mild, moderate or severe) and development of PDR was seen to increase with the duration of diabetes [7]. On a study done on 269 patients, Vitale et al. [8] found that the time of appearance of mild DR is not necessarily in coherence with appearance of PDR. Early detection and diagnosis of DR can prevent long term vision impairment effects by giving medical treatments to patients.

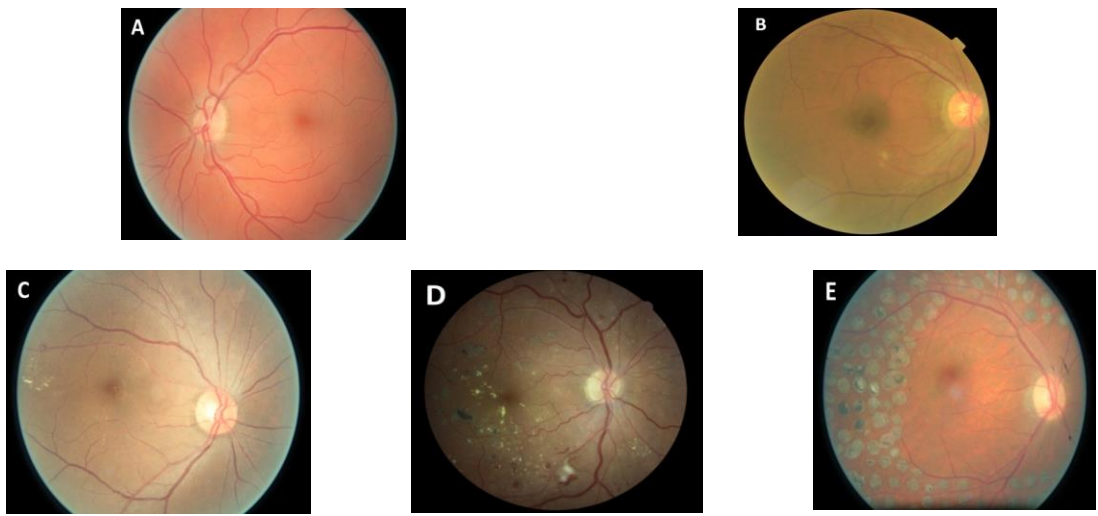

**Figure 1. Fundus images: (A) Normal (B) Mild (C) Moderate (D) Severe (E) Prolific DR**

### 1.2 Transfer Learning

Transfer learning is the reuse of deep learning models that are pre-trained on huge datasets such as subsets of the ImageNet project [9] to fit to a previously unseen dataset. Deep Learning (DL) is a subset of machine learning that takes advantage of the availability of huge training data and high computational power in solving problems such as classification and regression. Convolutional neural network (CNN) [10] is one example of DL architecture that are more suited to signals with multiple arrays such as images and videos [11]. DL depends on huge amount of training dataset to correctly work. The lack of enough annotated training data has been identified as one of the key challenges of applying DL in healthcare [12], [13]. Transfer learning, on deep convolutional neural networks, has gained traction due to the insufficiency of annotated training data in building models [14], [15]. Transfer learning can be used to reduce the training data required and minimize the DL training time.

In this paper, transfer learning on a deep CNN model that was pre-trained on a subset of the ImageNet's dataset was implemented. The pre-trained Inception-V3 model [16] was downloaded and a classifier was added to classify fundus images into healthy and unhealthy classes. Keras library was used to import the pre-trained model. A randomly selected subsample of the EyePacs DR Dataset which is hosted on Kaggle [17] platform is used for model training and testing.

This paper is structured as follows: DR classification research works related to ours are presented under Section 2. Section 3 presents the methods we used to classify fundus images and the test results are presented. And, Section 4 presents the final conclusions of our work.

## 2 Related Work

Feature extraction based classification and DL has been used to classify DR. In Acharya et al. [18] higher order spectra technique was used to extract features from 300 fundus images and fed to a support vector machine classifier; it classified the images into 5 classes with sensitivity of 82% and specificity of 88%. Different algorithms were developed to extract DR lesions such as blood vessels, exudates, and microaneurysms [19]. Exudates have been extracted for DR grading [20 - 24]. Support vector machine was used to classify the DIABETDB1 dataset into positive and negative classes using area and number of microaneurysms as features [25].

Feature extraction based classification methods need expert knowledge in order to detect the required features, and they also involve a time consuming process of feature selection, identification and extraction. Furthermore, DL based systems such as CNNs have been seen to outperform feature extraction based methods [26]. DL training for DR classification have been performed in two major categories: learning from scratch and transfer learning.

A convolutional neural network (CNN) was trained to classify a dataset of 128,175 fundus images into 2 classes, where the first class contains images with severity levels 0 and 1, and the second class contains levels 2, 3 and 4 [27]. In an operating cut point picked for high sensitivity, [27] had a sensitivity of 97.5% and specificity of 93.4% on the EyePACS-1 dataset which consists of 9963 images; it scored a sensitivity of 96.1% and a specificity of 93.9% on the Messidor-2 dataset; and in an evaluation cut point selected for high specificity, the sensitivity and specificity were 90.3% and 98.1% on the EyePACS-1, while 87% and 98.5% was scored on the Messidor-2, consecutively.

Using a training dataset of over 70,000 fundus images, Pratt et al. [28] trained a CNN using stochastic gradient descent algorithm to classify DR into 5 classes, and it achieved 95% specificity, 75% accuracy and 30% sensitivity. A DL model was trained from scratch on the MESSIDOR-2 dataset for the automatic detection of DR in [29], and a 96.8% sensitivity and 87% specificity was scored.

A CNN was trained from scratch to classify fundus images from the Kaggle dataset into referable and non-referable classes, and it scored a sensitivity of 96.2% and a specificity of 66.6% [30]. A dataset of 71896 fundus images was used to train a CNN DR classifier and resulted in a sensitivity of 90.5% and specificity of 91.6% [31]. A DL model was designed and trained on a dataset of 75137 fundus images and resulted in a sensitivity and specificity scores of 94% and 98%, respectively [32].

In order to avoid the time and resource consumed during DL, Mohammadian et al. [33] fine-tuned the Inception-V3 and Xception pre-trained models to classify the Kaggle dataset into two classes. After using data augmentation to balance the dataset, [33] reached at an accuracy score of 87.12% on the Inception-V3, and 74.49% on the Xception model. Wan et al. [34] implemented transfer learning and hyper parameter tuning on the pre-trained models AlexNet, VggNet-s, VggNet-16, VggNet-19, GoogleNet and ResNet using the Kaggle dataset and compared their performances. The highest accuracy score was that of VggNet-s model, which reached 95.68% when training with hyper-parameter tuning [34]. Transfer learning was used to work around the problem of insufficient training dataset in [35] for retinal vessel segmentation.

An Inception-V4 [36] model based DR classification scored higher sensitivity when compared with human expert graders on a 25,326 retinal images of patients with diabetes from Thailand [37].

Mansour [38] put to use the Kaggle dataset to train a deep convolutional neural network using transfer learning for feature extraction when building a computer aided diagnosis for DR. In Dutta et al. [39] 2000 fundus images were selected from the Kaggle dataset to train a shallow feed forward neural network, deep neural network and VggNet-16 model. On a test dataset of 300 images, the shallow neural network scored an accuracy of 42%, and the deep neural network scored 86.3% whiles the VggNet-16 scored 78.3% accuracy [39].

A training dataset of size 4476 was collected and labeled into 4 classes denpending on abnormalities and required treatment [40]; they resized input images into 600x600 and cut every image into four 300x300 images, and fed these images into separate pre-trained Inception-V3 models, which they called the Inception@4. After it was seen that accuracy result of the Inception@4 surpassed the VggNet and ResNet models, it was deployed on a web based DR classification system.

A multi cell, multi task convolutional neural network that uses a combination of cross entropy and mean square error was developed to classify images from the Kaggle dataset into 5 DR degrees [41]. A binary tree based multi-class VggNet classifier was trained on the Kaggle dataset in Adly et al. [42], and it scored an accuracy of 83.2%, sensitivity of 81.8% and specificity of 89.3% on a validation dataset of 6000 fundus images. By making use of SVMs with fully connected layers based on the VggNet-19 model, Mateen et al. [43] reached at an accuracy of 98.34% when classifying DR on the Kaggle dataset.

The Kaggle dataset [17], which contains 35126 labeled fundus images, has been exhaustively used for DL based classification of DR research purposes. In order to tackle the problem of medical training data insufficiency for DL, our model was trained on a subsample of 2500 fundus images and tested it on 5000 previously unseen fundus images. As per date, to the best of our knowledge, no attempt has been made to train and test a very deep neural network for DR classification using small training data.

## 3  Methodology

Figure 2 shows the process flow diagram of our DR classification system.

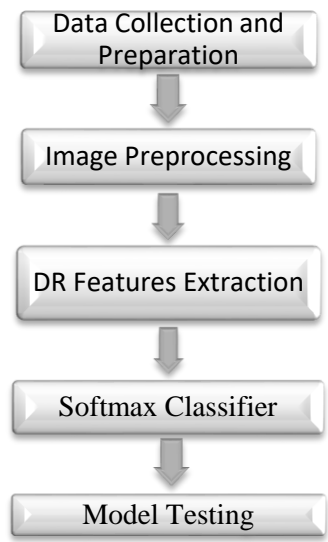

Figure 2. DR Detection System Diagram

### 3.1 Data Collection and Preparation

The Kaggle DR detection challenge dataset [17] contains color fundus images that are labeled 0,1,2,3 or 4 for normal, mild, moderate, severe and prolific DR, consecutively. We have reduced the DR classification into binary classes. A smaller subset, of size 2500 fundus images, of the publicly available EyePacs dataset that is uploaded on Kaggle DR Detection challenge was used for model training and testing. For model training, 1250 images were selected from the normal fundus images dataset for the healthy class, and 250 images were selected from each of the remaining four classes and put in the unhealthy class. For model testing, 1000 fundus images were selected from the normal dataset for healthy class and for the unhealthy class 1000 fundus images were selected from each of the four remaining classes.

### 3.2 Image Preprocessing

The fundus images were cropped from the input images to remove the black surrounding pixels using Opencv-Python. After cropping, the images were resized to 300x300. The local average was subtracted from each pixel [44]. A sample result of the image preprocessing done on the fundus images can be seen in Figure 3. Spyder IDE was used for image preprocessing.

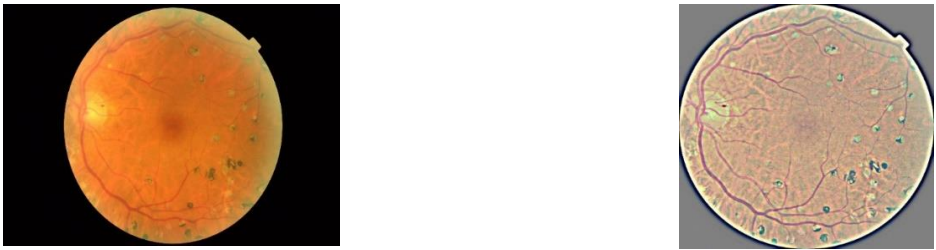

**Figure 3. Original fundus image (Left), cropped and preprocessed fundus image (Right)**

### 3.3 DR Features Extraction

The architecture of a deep CNN contains two basic parts: a convolutional and classifier part. While the convolutional part convolves with input data and extracts the separating characteristics of the different classes, the classifier part classifies the input data based on the collected characteristics. Many DL models that have scored high performance results when trained on subset of ImageNet dataset are made freely available in Keras. The convolutional part of these models is already trained to extract discriminating features from the ImageNet dataset.

In this work, we used the pre-trained convolutional part of Inception-V3 to extract features of fundus images.

Szegedy et al. [16] proposed Inception-V3 by adding convolution factorization and auxiliary classifiers in inception networks. Convolution factorization replaces bigger convolutions with multiple smaller convolutions in order to reduce network parameters, while auxiliary classifiers add regularization effect to the network. Inception networks are known for their use of inception modules. As can be seen in Figure 4, inception modules are a collection of different sized filters working on the same level of convolution on an input image. Inception modules help to extract features that differ in size among images of the same class. The concept of inception modules is helpful in DR classification because features of DR usually differ in size.

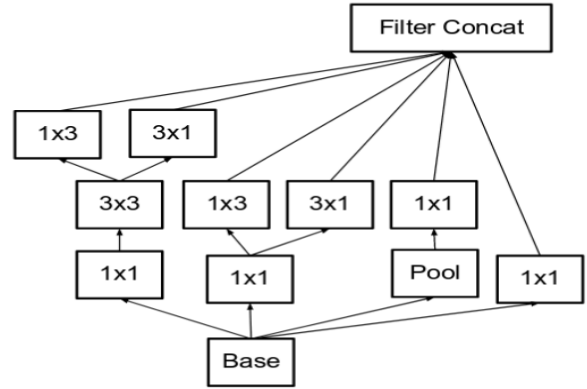

**Figure 4. Inception Module used in Inception-V3 [16]**

### 3.4 Classifier Training

We added a Relu activated layer followed by a Softmax classifier on top of the pre-trained models to classify the extracted features. Stochastic Gradient Descent (SGD) with an ascending learning rate of 0.0005 was used for training, and a Softmax function was used for the output layer. Cosine loss function was used to calculate error because it has shown better performance on small datasets [45].

### 3.5 Model Testing

We tested our model on a previously unseen dataset of 5000 fundus images. Our classifier model resulted in an accuracy of 90.9%, with a 3.94% loss. As can be seen in Table 1 our trained model has scored a higher accuracy than a model that employed a pre-trained Inception-V3 to classify the Kaggle dataset with data augmentation into binary classes [33].

**Table 1. Comparison of Results**

| Points of comparison | | Mohammadian et al. [33] | Proposed model |
|---|---|---|---|
| Training parameters | Size of training data | More than 35126 fundus images | 2500 fundus images |
| | Optimizer | ADAM | SGD |
| | Learning Rate | Not specified | Ascending rate of 0.0005 |
| | Loss function | Not specified | Cosine loss function |
| | Data Augmentation Used | Yes | No |
| Results | Accuracy | 87.12% | 90.9% |
| | Loss | Not specified | 3.94% |

Google Colaboratory, also known as Colabo, provides free of charge cloud GPU service for education and research purposes [46], [47]. Colabo was used for downloading the Inception-V3 pre-trained model, training and testing our model.

## 4 Conclusion

In this project, transfer learning is implemented to classify DR into 2 classes with a much reduced training data than other previous DR classification techniques employed. This was done to design a way to train a DL model that performs well on unseen data by efficiently learning from small dataset because training data is limited in healthcare. Our model has reached at an accuracy that is higher than other techniques that have used transfer learning on the whole Kaggle DR challenge dataset for binary classification. Our model has reached at a superior performance on account of the selected training algorithm, which is SGD with ascending learning rate, and the cosine loss function. Deep learning techniques that can learn from small dataset to categorize medical images should be utilized to classify DR, as this can be transferred to other medical image classification problems facing the challenge of insufficient training data. Experiments should be done to compare performances of other pre-trained deep convolutional networks in DR classification using small training data.